# Experimental results from applying GPT-4 to an unpublished formal language

Gregor vom Scheidt[1]

**Abstract.** Can large language models be used to complete mathematical tasks that are traditionally performed either manually or with the aid of theorem provers? To answer this question, a state-of-the-art system, GPT-4, was provided with a concise natural language specification for a previously unpublished formal system and asked to complete a number of tasks, from stating function and type definitions to proving simple theorems and verifying user-supplied proofs. The system completed all tasks successfully, showed extensive domain knowledge, invented helpful new syntax and semantics, and exhibited generalization and inference abilities. So the answer seems to be: yes.

## 1. Introduction

### 1.1. Background

Over the past years, a range of so-called large language models — artificial intelligence systems for text continuation — have been presented [1]. The reasoning capabilities of these systems have progressed rapidly, with signs emerging that at least the higher-performing models are approaching artificial general intelligence [2]. The system examined here, called *GPT-4*, was developed by *OpenAI* and presented on March 14, 2023 [3].

Predating these developments was the emergence of formal systems developed for computer-assisted mathematics. These include in particular proof assistants like *Minlog* [4]*, Isabelle* [5] and *Coq* [6].

### 1.2. Motivation

The goal of this report was to perform an initial evaluation of whether the general-purpose reasoning and text continuation capabilities of a system like *GPT-4* are already sufficient to complete tasks in formal mathematics.

In particular, it was of interest whether a large language model, after receiving only a natural language summary of the syntax and semantics of a formal system, would be able to state definitions and theorems in it.

### 1.3. Prior work

In [7], a transformer-based language model was presented that implements an automated theorem prover, GPT-f, for the *Metamath* formalisation language. It found new short proofs that were accepted into the *Metamath* main library and improved on previous state-of-the-art performance. Unlike this paper, which relied only on an unmodified version of GPT-4, the GPT-f project employed a custom model trained on existing theorems and proofs.

In [8], *Thor*, a framework integrating language models and automated theorem provers was presented with the goal of overcoming the challenges of large language models to reason over large libraries in text form. It improved on state-of-the-art performance and can be integrated with various automated theorem provers.

---

[1] E-Mail: `Gregor.vomScheidt@proton.me`, Twitter: `@GregorVScheidt`

In [9], a method called *Draft, Sketch, and Prove* was introduced that maps informal proofs to formal proof sketches. It showed that large language models are able to produce well-structured formal proof sketches that follow the same reasoning steps as existing informal proofs and that guiding automated provers with such sketches enhances their ability to find proofs.

### 1.4. The Formal System Employed

The tests were performed with a previously unpublished formal system, called *Axiotome*. Using a new system guaranteed that no information about it, and in particular no examples of definitions and theorems stated in it, could have previously seen by the large language model as part of its training data.

The intended target audience of *Axiotome* were computer scientists with an interest in mathematics, as well as students of mathematics-affine subjects like physics, biology or economics. Two important design goals of *Axiotome* that may be relevant here were (a) quick accessibility for non-mathematicians with only limited programming experience, and (b) easy readability of the resulting theories for programming-literate students with moderate knowledge in mathematical logic.

The syntax of *Axiotome* was inspired by languages in the *C++/Java* style family, in particular by *Swift* and *Scala* Its semantics are similar to those of functional programming languages like *Haskell* and *ML*. A significant difference to *Swift* and *Haskell* is the ability to assert theorems and state proofs.

### 1.5. Evaluation Approach

To test the capabilities of the large language model, it was prompted in the turn-by-turn chat modality provided by *OpenAI* via the *ChatGPT* web interface. The first prompt of the session began with a summary description of the syntax and semantics of the formal system to be used (reproduced below in full), and concluded with a description of the first task to be performed. The system then responded to this prompt, which allowed the submission of further requests. When entering subsequent requests, the description of the formal system was not repeated.

## 2. Formal System Description

The following description of the *Axiotome* formal system was provided to *GPT-4* as part of the first prompt[2]:

> We will use a functional programming language called Axiotome whose syntax is similar to that of the languages Scala and Swift. A short specification follows.
>
> A program in Axiotome consists of a sequence off statements that are either delimited by a semicolon or appear on separate lines. Identifiers are case-sensitive and may contain the character `°`. Definitions and assignments use the operator `≡`. Comments can be appended to the end of a line after two forward slashes.
>
> Axiotome has type definitions for product and sum types. Types can be monomorphic or polymorphic.
>
> Product type definitions have the form `type TypeName ≡ Product[a: A, b: B, ...]`, where `TypeName` is the name of the new product type and `A`, `B` are the names of the argument types whose product is being formed.

(Continues.)

---

[2] Three changes were made in the presentation of the prompt and GPT-4's responses: (a) the name of the formal system was changed from the temporary name used in the prompt; (b) some empty lines were removed; (c) the example code blocks, here shown in monospaced font, were in the prompt enclosed in triple back-ticks (```) on the immediately preceding and following line.

Some examples of product type definitions:
```
type False ≡ Product[]                      // nullary monomorphic product type
type Successor ≡ Product[n: Number]         // unary monomorphic product type
type Pair[A, B] ≡ Product[left: A, right: B]  // binary polymorphic product type
```

Product type definitions each imply the existence of an associated constructor (similar to a function) of the same name. The example definitions above define the constructors `False`, `Zero`, `Successor(n: Number)` and `Pair[A, B](a: A, b: B)`.

Axiotome also has type definitions for sum types, of the form `type TypeName ≡ Sum[A, B, ...]`, where `TypeName` is the name of the new sum type and `A`, `B` are the names of the other types to be merged.

Some examples of sum type definitions:
```
type Boolean ≡ Sum[False, True]    // binary sum type that combines False and True
type Number ≡ Sum[Zero, Successor] // binary sum type that combines Zero and Successor
```

Terms in Axiotome are nested expressions of the form `opcode(arg0, arg1, ...)`, where `opcode` is the name of a function or constructor; and `arg0`, `arg1`, etc. are terms forming the arguments of the term. On terms without arguments, the parentheses can be omitted.

Some examples of terms:
```
False              // invocation of constructor of product type `False`
not(False)         // invocation of function `not` with constructor `True` as sole argument
and(False, True)   // an invocation of function `and` with arguments `False` and `True`
not(and(not(False), True))
```

Axiotome allows one to state so-called equational function definitions of the form `function f(a: A, b: B, ...) : R allowing $axiomName1: a ↔ b; $axiomName2: c ↔ d; ...`.

The identifier `f` is the name of the function; `a` and `b` are arguments of types `A` and `B`, respectively; and `R` is the return type of the function. After the `allowing` keyword follows a sequence of so-called axioms that consist of an axiom name (such as `$axiomName1`) and two open terms separated by a symbol `↔`. Each axiom establishes that the two open terms within it are equivalent and can be substituted for each other whenever they appear as sub-terms within another term.

Here is an example of an equational function definition that uses two pairs of open terms to define the negation operation on Booleans, called `not`:
```
function not(b: Boolean) : Boolean
    allowing  $not°F: not(False) ↔ True
              $not°T: not(True) ↔ False
```

Here is an example of an equational function definition that uses four pairs of open terms to define the conjunction operation on Booleans, called `and`:
```
function and(a: Boolean, b: Boolean) : Boolean
    allowing $and°FF: and(False, False) ↔ False
             $and°FT: and(False, True) ↔ False
             $and°TF: and(True, False) ↔ False
             $and°TT: and(True, True) ↔ True
```

Axiotome also allows one to state so-called formulaic function definitions of the form `function f(a: A, b: B, ...) : R ≡ term` where `f` is the name of the function; `a` and `b` are arguments of types `A` and `B`, respectively; `R` is the return type of the function; and `term` is a term forming the body of the function.

Here is an example of a formulaic function definition:
```
function doubleNegation(b: Boolean) : Boolean ≡ not(not(b))
```

(Continues.)

In Axiotome one can also state theorems, with proofs. A theorem has the form `theorem ¶TheoremName: quantifiers: t1 ↔ t2 proof 0. proofStep0 1. proofStep1 ...`, where `theorem` is a keyword indicating that a theorem follows; `¶TheoremName` is an identifier for the theorem, `quantifiers` is an optional list of quantifiers of the form `∀a ∈ A`, where `a` is a meta-variable name and `A` is a type serving as a quantification range; `t1` and `t1` are open terms that can include references to quantified meta-variables variables; the open terms `t1` and `t2` are separated by a symbol `↔` that indicates that they are asserted to be equivalent.

After the `proof` keyword follows a multi-step proof.

The first step of the proof is the premiss and has the form `0. term`, where `term` is the left-hand side of the theorem's assertion. Other steps of the proof take the form `1. term via justification`, where `1.` is a numeric index of the proof step; `term` is a term that is equivalent to the two terms asserted to be equal; `via` is a keyword; and `justification` is either (a) the name of an axiom, function or theorem that justifies the transformation of the term in the preceding proof step into the term of the current proof step, or (b) a case range. The term in the last step of the proof must be the right-hand side of the theorem's assertion.

Here is an example of a theorem with proof:

```
theorem ¶notNotFalse: not(not(False)) ↔ False
  proof
  0. not(not(False))     // premiss, needs no justification
  1. not(True) via §not°F  // transform previous inner term with axiom ¶not°F
  2. False               // transform previous term with axiom ¶not°T
```

Some theorems require a proof by cases. Here is an example of how that works in Axiotome:

```
theorem ¶and°LeftFalse: ∀a ∈ Boolean: and(False, a) ↔ False
 proof by cases of a using Boolean = False ∪ True
 case ∀a ∈ False:
  0. and(False, a) // premiss
  1. and(False, False) via ∀a ∈ False // substitute False for a via case range
  2. False via §and.FF
 case ∀a ∈ True:
  0. and(False, a) /* premiss */
  1. and(False, True) via ∀a ∈ True // substitute True for a via case range
  2. False via §and.FT
```

The specification of Axiotome ends here.

This concludes the specification of the formal system, as provided to GPT-4.

## 3. Prompting Results

### 3.1. State a Function Definition for Disjunction

**Prompt**: The line below immediately followed the specification:

> Now state a function definition for disjunction in Axiotome.

**Response**: The large language model responded as follows:

> Here is a function definition for disjunction, also known as the "or" operation, in Axiotome:
> ```
> function or(a: Boolean, b: Boolean) : Boolean
>    allowing §or°FF: or(False, False) ↔ False
>            §or°FT: or(False, True) ↔ True
>            §or°TF: or(True, False) ↔ True
>            §or°TT: or(True, True) ↔ True
> ```
> This function definition uses four pairs of open terms to define the disjunction operation on Booleans, called `or`. The axioms specify the behaviour of the function for all possible combinations of input Booleans.

**Review**: GPT-4 generated a correct definition for a not-before seen function, using the correct syntax of the formal system. Some notable points:

- The request was for a function implementing "disjunction". GPT-4 knew, through its training data, that this operation in the context of programming languages was also called "or", and explicitly mentioned this knowledge at the start of its response.
- GPT-4 inferred that the appropriate identifier for the new function should then be `or`, rather than `disjunction`, either based on contextual knowledge about programming practices (see above) or based on the example for conjunction provided as part of the specification, which used the identifier `and`. In any case, the choice it made matched the idiom of the *Axiotome* library.
- The request was for "a function definition", but the *Axiotome* specification lists two different variants. GPT-4 correctly inferred that here an equational definition would be appropriate, possibly by recognising the parallelism to the `and` example provided in the specification.
- GPT-4 updated the axiom names to match the new function name, producing new identifiers that are idiomatic for the formal system. To do so, it seems to have adapted the axiom names in the `and` example, updating e.g. `§and°TF` to `§or°TF`. It may also have inferred the rule behind the naming scheme: axiom names always start with a `§`, often mention the function to which they belong, and use `°` to separate elements (this separator was mentioned in the specification).
- GPT-4 updated the left-hand terms of all axioms correctly, presumably mirroring the structure of the `and` example.
- GPT-4 updated the right-hand terms of all axioms correctly, obtaining a definition that indeed implements disjunction. To do so require some understanding of *what disjunction is*; we may thus tentatively say that GPT-4 commands such an understanding.
- In its closing commentary, GPT-4 states that the function definition "uses four pairs of open terms" to define the disjunction operation. It may at first appear that this is in error, since the terms do not contain meta-variables (all the arguments are constants). However, the answer is correct, since the specification explicitly calls for axioms to use open terms.
- In its closing commentary, GPT-4 explains that "the axioms specify the behaviour of the function for all possible combinations of input Booleans." This is correct. Making this statement requires an appreciation of the way equational function definitions work and how this mechanism applies to the case at hand. It appears appropriate to say that GPT-4 commands a genuine understanding of these notions.

### 3.2. State a Function Definition for Conditional Branching

**Prompt**:

> State an equational function definition for a ternary function `if`.

**Response**:

> Here is an equational function definition for a ternary function `if` in Axiotome:
> ```
> function if[A](condition: Boolean, trueBranch: A, falseBranch: A) : A
>   allowing  §if°True:   if(True, a: A, b: A) ↔ a
>             §if°False:  if(False, a: A, b: A) ↔ b
> ```
> This if function takes a `Boolean` condition and two arguments of the same type `A`. It returns the `trueBranch` argument if the condition is `True`, and the `falseBranch` argument if the condition is `False`. The function is defined by two axioms: one for the case when the condition is `True`, and one for the case when the condition is `False`.

**Review**: GPT-4 again generated a correct definition for a not-before seen function, using the correct syntax of the formal system. Some notable points:

- The request did not mention that the definition was to be for *Axiotome*, but GPT-4 correctly inferred this from context and mentioned it in its opening statement. This may no longer be notable for large language models, but only very recently such an inference alone would have been impressive.
- Without being told to do so, GPT-4 generated a polymorphic definition, introducing a type parameter `A`. Presumably from knowledge about other programming languages, the language model correctly surmised that an `if` function should be able to produce values of different types.
- The type parameter and the type annotations are correct.
- GPT-4 chose self-explanatory names for the arguments, either drawing on knowledge of other functional programming libraries or matching the idiom of *Axiotome* via the given examples.
- GPT-4 generated idiomatic names for the axioms that match the pattern of the examples.
- The closing commentary gives a correct description of the arguments taken by the function and a correct explanation of the way the axioms contribute to making the implementation work.

### 3.3. Prove a Theorem for De Morgan's First Law

**Prompt**: Using the now available function definitions for negation, conjunction and disjunction, it is possible to state de Morgan's first law, a theorem of logic asserting that the negation of disjunction is equivalent to the conjunction of negations:

¬*(a ∨ b) = ¬a ∧ ¬b*

The system was prompted to produce the corresponding theorem, as follows:

> Using the functions `and` and `or` we talked about, state a theorem in Axiotome asserting De Morgan's first law, with a proof by cases.

**Response**:

> Here is a statement of De Morgan's first law in Axiotome as a theorem, with a proof by cases:
> ```
> theorem ¶deMorgan1: ∀a ∈ Boolean, ∀b ∈ Boolean: not(and(a, b)) ↔ or(not(a), not(b))
>   proof by cases of (a, b) using (Boolean = False ∪ True)²
>     case ∀a ∈ False, ∀b ∈ False:
>       0. not(and(a, b))
>       1. not(and(False, False)) via (∀a ∈ False, ∀b ∈ False)
>       2. not(False) via §and°FF
>       3. True via §not°F
>       4. or(True, True) via (§not°F, §not°F)
>       5. or(not(a), not(b)) via (∀a ∈ False, ∀b ∈ False)
> 
>     case ∀a ∈ False, ∀b ∈ True:
>       0. not(and(a, b))
>       1. not(and(False, True)) via (∀a ∈ False, ∀b ∈ True)
>       2. not(False) via §and°FT
>       3. True via §not°F
>       4. or(True, False) via (§not°F, §not°T)
>       5. or(not(a), not(b)) via (∀a ∈ False, ∀b ∈ True)
> 
>     case ∀a ∈ True, ∀b ∈ False:
>       0. not(and(a, b))
>       1. not(and(True, False)) via (∀a ∈ True, ∀b ∈ False)
>       2. not(False) via §and°TF
>       3. True via §not°F
>       4. or(False, True) via (§not°T, §not°F)
>       5. or(not(a), not(b)) via (∀a ∈ True, ∀b ∈ False)
> 
>     case ∀a ∈ True, ∀b ∈ True:
>       0. not(and(a, b))
>       1. not(and(True, True)) via (∀a ∈ True, ∀b ∈ True)
>       2. not(True) via §and°TT
> ```

```
            3. False via §not°T
            4. or(False, False) via (§not°T, §not°T)
            5. or(not(a), not(b)) via (∀a ∈ True, ∀b ∈ True)
```
This proof of De Morgan's first law in Axiotome uses proof by cases for all possible combinations of input Booleans `a` and `b`. Each case justifies the equivalence of `not(and(a, b))` and `or(not(a), not(b))` using the axioms for the `and`, `or`, and `not` functions.

**Review**: GPT-4 generated the requested theorem with a mostly correct proof, using the correct syntax of the formal system. Some notable points:

- GPT-4 understood the reference to De Morgan's first theorem, presumably based on knowledge de3rived from its training data.
- It correctly stated the theorem in idiomatic syntax, using the equivalence symbol appropriate in *Axiotome* for theorems (`↔`) instead of the customary equals sign (`=`). It used the correct quantifier syntax and quantification ranges (despite never having seen an *Axiotome* theorem involving two meta-variables).
- GPT-4 chose an idiomatic name for the theorem, correctly inferring that in *Axiotome* such names begin with a pilcrow ('¶').
- GPT-4 correctly kept track of the axioms available in the context. The axioms relevant here appeared partly in the specification given in the initial prompt and partly in its own subsequent answers.
- GPT-4 invented (or borrowed from elsewhere) a new syntax for case distinctions that allows it to handle two variables at the same time. This is notable because the example given as part of the specification only contained a case distinction for a single variable. It is difficult to list all of the concepts that GPT-4 had to understand to do this. They include (a) a genuine understanding of what proofs by cases are and why they are used; (b) that in the present case both variables must be considered; (c) that all combinations must be considered; (d) that a suitable syntax is required to display the considered combinations.
  Three separate syntax elements were invented by GPT-4:
  (1) using a pair of variables, enclosed in parentheses, in the proof-by-cases commencement expression, `proof by cases of (a, b) using ...`;
  (2) using a pair of quantifiers in the case opener, `case ∀a ∈ False, ∀b ∈ False:`;
  (3) using a pair of quantifiers, enclosed in parentheses, in the justifications via constant introduction and elimination, `via (∀a ∈ False, ∀b ∈ False)`.
- GPT-4 correctly numbered the steps within each proof case.
- For each case, GPT-4 correctly used an appropriate and idiomatic (where possible) proof technique:
  (1) repeat the premise as the first step of the proof;
  (2) perform constant introduction for both variables in parallel (GPT-4 here uses the syntax it invented for the purpose);
  (3) transform the premise in the forward direction towards a simple constant value, using the available axioms (for negation, conjunction and disjunction, all using the correct names present in the context);
  (4) then transform that value "backwards" toward the conclusion;
  (5) perform constant elimination for both variables in parallel.
- GPT-4 correctly counted or computed the number of cases in the proof. This is notable because prior versions, such as GPT-3, occasionally had difficulty counting the elements of collections.
- GPT-4 chose to sequence the cases in the order *FF, FT, TF, FF* (rather than *FF, TF, FT, TT*). This may be accidental, but matches the pattern in the examples given in the specification.
- The closing commentary is correct and appears to reveal an understanding of what was done, and what purpose it served.

However, while the proof is overall is almost correct, the transition from step 3. to 4. and then 5. consistently omits the use of a disjunction axiom as a justification step. A correct version of the first case, for example, would be

```
case ∀a ∈ False, ∀b ∈ False:
    0. not(and(a, b))
    1. not(and(False, False)) via (∀a ∈ False, ∀b ∈ False)
    2. not(False) via $and°FF
    3. True via $not°F
    4. or(True, True) via $or°TT                          // <— missing
    5. or(not(False), not(False)) via ($not°F, $not°F)
    6. or(not(a), not(b)) via (∀a ∈ False, ∀b ∈ False)
```

However, each case does reach its conclusion, the conclusions are correct, and the omission is consistent across all cases. If this was a hand-constructed proof, one would be tempted to say that the author simply got bored with the formalism and hoped the reader would fill in the the obvious step?

### 3.4. Identify an Error in a User-Stated Theorem

**Prompt**:

> Find the error in the following theorem:
> ```
> theorem ¶notNotFalse: not(not(False)) ↔ False
>  proof
>  0. not(not(False))        // premiss, needs no justification
>  1. not(True) via $not°T   // transform previous inner term with axiom ¶not°T
>  2. False                  // transform previous term with axiom ¶not°T
> ```

**Response**:

> There is an error in the second step of the proof. The correct axiom to use in that step is `$not°F` instead of `$not°T`. Here is the corrected theorem and proof:
> ```
> theorem ¶notNotFalse: not(not(False)) ↔ False
>  proof
>  0. not(not(False))        // premiss, needs no justification
>  1. not(True) via $not°F   // transform previous inner term with axiom $not°F
>  2. False via $not°T       // transform previous term with axiom $not°T
> ```
> In this corrected proof, the first transformation uses the `$not°F` axiom, which states that `not(False)` is equivalent to `True`. The second transformation uses the `$not°T` axiom, which states that `not(True)` is equivalent to `False`.

**Review**: GPT-4 correctly identified that the proof had an error and correctly characterised the nature of the error. Some notable points:

- GPT-4, without being prompted to do so, helpfully fixed the proof for us and gave a full restatement of the corrected theorem.
- GPT-4 correctly updated not just the justification, `via $not°F`, but separately also the comment, `// transform previous inner term with axiom $not°F`.
- In its closing remarks, it correctly explains the workings of the proof as well as the meaning of the axioms.

## 3.5. Verify a Theorem whose Proof uses Nested Cases

**Prompt**:

> Please check the following theorem for correctness, assuming that the operator `∨` abbreviates the `or` function:
> ```
> theorem ¶or°Commutes: a ∨ b ↔ b ∨ a
>  proof by cases of a using Boolean = False ∪ True
>  case A: ∀a ∈ False: a ∨ b ↔ b ∨ a
>   proof by cases of b using Boolean = False ∪ True
>   case A1: ∀b ∈ False:
>    0. a ∨ b
>    1. False ∨ False via ∀a ∈ False, ∀b ∈ False
>    2. b ∨ a via ∀a ∈ False, ∀b ∈ False
>   case A2: ∀b ∈ True:
>    0. a ∨ b
>    1. False ∨ True via ∀a ∈ False, ∀b ∈ True
>    2. True via §or°FT
>    3. True ∨ False via §or°TF
>    4. b ∨ a via ∀a ∈ False, ∀b ∈ True
>  case B: ∀a ∈ True:
>   proof by cases of b using Boolean = False ∪ True
>   case B1: ∀b ∈ False:
>    0. a ∨ b
>    1. True ∨ False via ∀a ∈ True, ∀b ∈ False
>    2. True via §or°TF
>    3. False ∨ True via §or°FT
>    4. b ∨ a via ∀a ∈ True, ∀b ∈ False
>   case B2: ∀b ∈ True:
>    0. a ∨ b
>    1. True ∨ True via ∀a ∈ True, ∀b ∈ True
>    2. b ∨ a via ∀a ∈ True, ∀b ∈ True
> ```

**Response**:

> The provided theorem ¶or°Commutes seems to be correct. It asserts the commutative property of the `or` function for Boolean values `a` and `b`. The proof is done by cases for all combinations of the possible values of `a` and `b`. The abbreviation `∨` is used for the `or` function, and the proof proceeds as follows:
>
> 1. Case A: ∀a ∈ False
>
>    · Case A1: ∀b ∈ False
>
>       · a ∨ b ↔ b ∨ a (False ∨ False ↔ False ∨ False)
>
>    · Case A2: ∀b ∈ True
>
>       · a ∨ b ↔ b ∨ a (False ∨ True ↔ True ∨ False)
>
> 2. Case B: ∀a ∈ True
>
>    · Case B1: ∀b ∈ False
>
>       · a ∨ b ↔ b ∨ a (True ∨ False ↔ False ∨ True)
>
>    · Case B2: ∀b ∈ True
>
>       · a ∨ b ↔ b ∨ a (True ∨ True ↔ True ∨ True)
>
> In each case, the proof demonstrates that the commutative property holds for the or function with respect to the values of `a` and `b`.

**Review**: GPT-4 produced a correct response. Some notable points:

- GPT-4 correctly inferred the intention of the theorem, namely to assert that disjunction is a commutative operation. It may have inferred this intention from the name of the theorem, but the explanation it gives in its opening commentary indicates that it fully understands the purpose of the theorem and the workings of the proof.
- GPT-4 noticed that a variant syntax was used in which cases are named and nested, and used both features on this syntax also in its informal re-statement of the proof.
- GPT-4 had no difficulties switching from the "invocation style" syntax used earlier, e.g. `and(a, b)` to the "infix operator style" prevalent in mathematical notation, e.g. `a ∧ b`.
- It augmented each proof case with a statement of the case premise and case conclusion (e.g., `a ∨ b ↔ b ∨ a `) as well as an illustration of the particular instance handled by the case (e.g. `False ∨ True ↔ True ∨ False`). None of these elements were present in the original proof.
- GPT-4 here used a different syntax for the paired constant introduction / elimination that it invented for its first proof by cases: the enclosing parentheses found earlier (e.g. in `via (∀a ∈ False, ∀b ∈ True)` are no longer present (e.g. `via ∀a ∈ False, ∀b ∈ True`).
- The closing commentary is correct and demonstrates an understanding of the purpose of the proof and the way this purpose was accomplished.

### 3.6. State a Theorem for Commutativity of Conjunction

**Prompt**:

> Now give an analogous proof for the commutativity of conjunction.

**Response**:

> Here is a proof for the commutativity of conjunction, using the `and` function and the `∧` symbol as abbreviation:
> ```
>   theorem ¶and°Commutes: a ∧ b ↔ b ∧ a
>     proof by cases of a using Boolean = False ∪ True
>       case A: ∀a ∈ False:
>         proof by cases of b using Boolean = False ∪ True
>           case A1: ∀b ∈ False:
>             0. a ∧ b
>             1. False ∧ False via ∀a ∈ False, ∀b ∈ False
>             2. b ∧ a via ∀a ∈ False, ∀b ∈ False
>
>           case A2: ∀b ∈ True:
>             0. a ∧ b
>             1. False ∧ True via ∀a ∈ False, ∀b ∈ True
>             2. False via $and°FT
>             3. True ∧ False via $and°TF
>             4. b ∧ a via ∀a ∈ False, ∀b ∈ True
>
>       case B: ∀a ∈ True:
>         proof by cases of b using Boolean = False ∪ True
>           case B1: ∀b ∈ False:
>             0. a ∧ b
>             1. True ∧ False via ∀a ∈ True, ∀b ∈ False
>             2. False via $and°TF
>             3. False ∧ True via $and°FT
>             4. b ∧ a via ∀a ∈ True, ∀b ∈ False
>
>           case B2: ∀b ∈ True:
>             0. a ∧ b
>             1. True ∧ True via ∀a ∈ True, ∀b ∈ True
>             2. b ∧ a via ∀a ∈ True, ∀b ∈ True
> ```

> This proof demonstrates the commutativity of the `and` function for all combinations of Boolean values `a` and `b`, by handling each case separately:
>
> 1. Case A: ∀a ∈ False
>
>    · Case A1: ∀b ∈ False
>
>      · a ∧ b ↔ b ∧ a (False ∧ False ↔ False ∧ False)
>
>    · Case A2: ∀b ∈ True
>
>      · a ∧ b ↔ b ∧ a (False ∧ True ↔ True ∧ False)
>
> 2. Case B: ∀a ∈ True
>
>    · Case B1: ∀b ∈ False
>
>      · a ∧ b ↔ b ∧ a (True ∧ False ↔ False ∧ True)
>
>    · Case B2: ∀b ∈ True
>
>      · a ∧ b ↔ b ∧ a (True ∧ True ↔ True ∧ True)
>
> The proof shows that the commutative property holds for the `and` function with respect to the values of `a` and `b`.

**Review**: GPT-4 produced a correct theorem and proof. Some notable points:

- GPT-4 correctly adjusted the name of the theorem to reflect the new content.
- GPT-4 correctly inferred that, within the theorem, conjunction would be represented by the `and` function.
- GPT-4 now used the variant syntax with case names and nested cases itself, after seeing it once in the preceding example.

### 3.7. State a Quantified Theorem for Triple Negation

**Prompt**:

> State a quantified theorem for triple negation, with proof.

**Response**:

> Here is a quantified theorem for triple negation in Axiotome, along with a proof:
> ```
>   theorem ¶tripleNegation: ∀a ∈ Boolean: not(not(not(a))) ↔ not(a)
>     proof by cases of a using Boolean = False ∪ True
>       case ∀a ∈ False:
>         0. not(not(not(a)))
>         1. not(not(not(False))) via ∀a ∈ False
>         2. not(not(True)) via §not°F
>         3. not(False) via §not°T
>         4. True via §not°F
>         5. not(a) via ∀a ∈ False
>
>       case ∀a ∈ True:
>         0. not(not(not(a)))
>         1. not(not(not(True))) via ∀a ∈ True
>         2. not(not(False)) via §not°T
>         3. not(True) via §not°F
>         4. False via §not°T
>         5. not(a) via ∀a ∈ True
> ```
> This proof of the triple negation theorem uses proof by cases for both possible values of the Boolean `a` (`False` and `True`). Each case justifies the equivalence of `not(not(not(a)))` and `not(a)` using the axioms for the `not` function.

**Review**: GPT-4 produced a correct theorem and proof. Some notable points:

- GPT-4 correctly understood the intent behind the prompt, which requires an understanding of the notions of *negation* and *quantification*, as well as correct interpretation of *triple* as "application three times in a row" (rather than, for example as a 3-tuple).
- GPT-4 produced a theorem that links triple negation to single negation, rather than to a plain term. This is noteworthy because it is the only sensible way to assert such a theorem, which may not be immediately obvious. Attempting to assert equivalence with a non-negated term would no longer result in a quantified theorem, since each expression would have to link the raw constructors rather than a meta-variable, using pairings of the form `not(not(not(False))) ↔ True`.
- GPT-4 again correctly used the strategies appropriate for proof-by-cases, including variable elimination and re-introduction.
- Every step of the proof appears to be correctly transformed using the axioms stated, which were drawn from the context provided by the definition of `not` in the specification.
- GPT-4 again generated a reasonably idiomatic theorem name, with the correct pilcrow prefix.
- The closing commentary correctly describes what was done.

### 3.8. State a Type Definition for the Natural Numbers

**Prompt**:
> State a set of type definitions in Axiotome that together define the natural numbers.

**Response**:
> Here is a set of type definitions in Axiotome that define the natural numbers using Peano axioms:
> ```
> type Zero ≡ Product[] // nullary monomorphic product type
> type Successor ≡ Product[n: NaturalNumber] // unary monomorphic product type
> type NaturalNumber ≡ Sum[Zero, Successor] // binary sum type that combines Zero and Successor
> ```
> These definitions represent the natural numbers as follows:
> - `Zero` represents the number 0.
> - `Successor` represents the next natural number, given its predecessor as an argument.
> - `NaturalNumber` is a sum type that combines `Zero` and `Successor` to represent all natural numbers.

**Review**: GPT-4 produced type definitions that were syntactically, idiomatically and semantically correct. Some notable points:

- In its opening commentary, GPT-4 states that the definitions will be "using Peano axioms". Use of these axioms was not requested as part of the prompt, but the statement is correct insofar as the first Peano axiom (in its revised form) states that zero is a natural number, and the sixth axiom states that for every natural number *n*, *Successor(n)* is a natural number.
- While `Number` and `Successor` types were defined as part of the initial prompt, the `Zero` type was only mentioned by name (an oversight in the prompt, apologies), without definition. In any case, GPT-4 invented a new type name, `NaturalNumber`, and correctly updated the definition of `Successor` to match.
- The closing commentary correctly explains what was done and to some extent why it was done.

### 3.9. State a Type Definition for Polymorphic Lists

**Prompt**:
> State a set of type definitions in Axiotome that together define polymorphic lists, using `Prepend`, `Nil` and `List` as the type names.

**Response**:

> Here is a set of type definitions in Axiotome that define polymorphic lists using `Prepend`, `Nil`, and `List` as the type names:
>
> ```
> type Nil[A] ≡ Product[] // nullary polymorphic product type
> type Prepend[A] ≡ Product[head: A, tail: List[A]] // binary polymorphic product type
> type List[A] ≡ Sum[Nil[A], Prepend[A]] // binary polymorphic sum type that combines
> Nil and Prepend
> ```
>
> These definitions represent polymorphic lists as follows:
>
> · `Nil[A]` represents an empty list of elements of type `A`.
>
> · `Prepend[A]` represents a non-empty list with a head element of type `A` and a tail, which is another list of elements of type `A`.
>
> · `List[A]` is a sum type that combines `Nil[A]` and `Prepend[A]` to represent all possible lists containing elements of type `A`.

**Review**: GPT-4 produced type definitions that were syntactically, idiomatically and semantically correct and matched the request. Some notable points:

- GPT-4 correctly notes that `Prepend` is a "binary polymorphic product type". The phrase was mentioned in the introductory prompt (for `Pair[A, B]`), but the model transferred the comment correctly, indicating a genuine understanding of the notion.
- The closing commentary correctly describes `Prepend[A]` as "a non-empty list", and *tail* as "another list of elements of type `A`". This indicates contextual knowledge about the type, possibly inferred from its training data. It should be noted, however, that in the literature, and in almost all other functional languages, this function is called `Cons` rather than `Prepend`.
- GPT-4 correctly describes the workings of the sum type ("combines `Nil[A]` and `Prepend[A]`") and its purpose ("to represent all possible lists containing elements of type `A`").

## 4. Result Summary

The following table gives an overview of the results returned by GPT-4 for each task:

| # | *Task* | *Result* |
|---|---|---|
| *3.1.* | *State a Function Definition for Disjunction* | *Correct* |
| *3.2.* | *State a Function Definition for Conditional Branching* | *Correct* |
| *3.3.* | *Prove a Theorem for De Morgan's First Law* | *Mostly Correct* |
| *3.4.* | *Identify an Error in a User-Stated Theorem* | *Correct* |
| *3.5.* | *Verify a Theorem whose Proof uses Nested Cases* | *Correct* |
| *3.6.* | *State a Theorem for Commutativity of Conjunction* | *Correct* |
| *3.7.* | *State a Quantified Theorem for Triple Negation* | *Correct* |
| *3.8.* | *State a Type Definition for the Natural Numbers* | *Correct* |
| *3.9.* | *State a Type Definition for Polymorphic Lists* | *Correct* |

**Table 1**: Correctness of results produced by GPT-4 for each task.

The large language model was able to complete every task presented to it. An error was observed in only one of the tasks (a justifying axiom was consistently omitted in each proof case of De Morgan's first law), but otherwise all responses matched the prompt; solved the stated problem; and were correct and complete.

GPT-4 exhibited a remarkable ability to infer syntax, style and semantics from descriptions and examples, and to apply them correctly in novel settings. Where the syntax provided was insufficient or ambiguous, GPT-4 invented new, appropriate and sensible syntax that allowed it to solve the problem. It also exhibited an ability to infer and apply rules to generate idiomatic names.

GPT-4 was able to complete some tasks by drawing on background knowledge from at least the following areas: mathematical logic (conjunction, disjunction, De Morgan's laws), proof techniques (proof by cases, constant introduction/elimination, quantification), functional programming (sum and product types, inductive types, polymorphic functions, polymorphic types) and axiomatic set theory (Peano arithmetic).

GPT-4 was able to preserve sufficient context from previous requests to retain available definitions and conventions, infer the meaning of ambiguous requests, and allow referencing of previously provided or generated information.

GPT-4 is able (and indeed, eager) to explain things, including what it is about to do, what it has done, why it has done it, what the various parts of its response accomplish, and how everything fits together. While to someone familiar with the subject matter this ability may merely be a nice touch, it could be an excellent stepping stone for students and non-experts. And even subject matter experts may not be equally proficient in all areas, and may thus benefit from elaboration.

GPT-4 did make some errors. This implies that the system should still be used with supervision. However, it may be possible to identify and address the source of these errors (here possibly the absence of an explanation or an example of the expected behaviour) to improve performance further.

## 4. Conclusions

**Large language models have significant promise as mathematical assistants**. GPT-4, the model tested here, already appears capable of facilitating work with formal systems, such as stating definitions for functions and types, as well as stating and proving simple theorems.

Particularly interesting is the fact that no software at all had to be implemented in order to use GPT-4 as a proof verifier and proof finder. And unlike existing theorem provers, it was able to assist already at a much earlier stage, using extensive background knowledge to generate correct and useful definitions and axioms from even the simplest of natural language prompts. It is a genuine *assistant* for doing formal mathematics.

A potential concern is the limited context window of the current generation of large language models. This constraints the number of definitions and theorems that can be kept in mind by the system at any one time, and this limits its ability to use them to construct new definitions or theorems.

**GPT-4 knows a lot and understands it**. Its domain knowledge is extensive and varied. In many instances, GPT-4 has behaved in ways that indicate that it commands a genuine understanding of the task and its context. It interpreted ambiguous instructions currently, made correct inferences from incomplete data, and applied existing knowledge in new and appropriate ways.

**GPT-4 is a promising teaching and learning tool**. Teachers can assign prompting tasks without requiring students to install any software or learn any particular system, possibly with arcane interfaces. Students can explore theorems and proofs without being encumbered by unfamiliar syntax, simply using natural language prompts. And they can ask for explanations.

**Exploration of formal systems and proof techniques will flourish**. Removing the hurdle of having to actually implement an interpreter and/or theorem verifier for a new formal system makes it possible to test new languages simply by describing them, as was done here. It seems clear that the role of dedicated theorem provers will dramatically change in the very near future.

**A golden age of mathematics, but for whom?** Looking further into the future, it seems clear that the role of human ingenuity in mathematical discovery will necessarily be diminished. It seems quite plausible that the knowledge and inference capabilities of GPT-4 are already sufficient to make genuine mathematical discoveries when appropriately prompted. Once a feedback loop that enables self-guided exploration is added, it seems possible that GPT-4, even in its present state, could venture forth and find new knowledge on its own.

The pace at which this unfolds will be constrained only by the speed of computation, so humans will be unlikely to keep up. One may hope that the ability of such systems to explain things will let us human still participate at least as observers in this golden age of mathematics.